\documentclass[10pt,twocolumn,letterpaper]{article}

\usepackage[pagenumbers]{cvpr} 

\usepackage{graphicx}
\usepackage{amsmath}
\usepackage{amssymb}
\usepackage{booktabs}
\usepackage{xcolor}

\usepackage[pagebackref,breaklinks,colorlinks]{hyperref}

\usepackage[capitalize]{cleveref}
\crefname{section}{Sec.}{Secs.}
\Crefname{section}{Section}{Sections}
\Crefname{table}{Table}{Tables}
\crefname{table}{Tab.}{Tabs.}

\usepackage{multirow}

\usepackage{pifont}
\newcommand{\cmark}{\ding{51}}%
\newcommand{\xmark}{\ding{55}}%
\usepackage{xcolor}

\newcommand\Tstrut{\rule{0pt}{2ex}}         
\newcommand\Bstrut{\rule[-1ex]{0pt}{0pt}}   

\def\cvprPaperID{7660} 
\def\confName{CVPR}
\def\confYear{2023}

\begin{document}

\hbadness=2000000000
\vbadness=2000000000
\hfuzz=100pt

\setlength{\abovedisplayskip}{1pt}
\setlength{\belowdisplayskip}{1pt}
\setlength{\floatsep}{6pt plus 1.0pt minus 1.0pt}
\setlength{\intextsep}{6pt plus 1.0pt minus 1.0pt}
\setlength{\textfloatsep}{6pt plus 1.0pt minus 1.0pt}

\title{Leveraging Endo- and Exo-Temporal Regularization for\\ Black-box Video Domain Adaptation}

\author{Yuecong Xu\\
Institute for Infocomm Research, A*STAR, Singapore\\
{\tt\small xuyu0014@e.ntu.edu.sg}
\and
Jianfei Yang, Haozhi Cao, Lihua Xie\\
School of Electrical and Electronic Engineering, NTU, Singapore\\
{\tt\small \{yang0478, haozhi001\}@e.ntu.edu.sg, elhxie@ntu.edu.sg}
\and
Min Wu, Xiaoli Li, Zhenghua Chen\\
Institute for Infocomm Research, A*STAR, Singapore\\
{\tt\small \{wumin,xlli\}@i2r.a-star.edu.sg, chen0832@e.ntu.edu.sg}
}

\maketitle

\begin{abstract} 
   To enable video models to be applied seamlessly across video tasks in different environments, various Video Unsupervised Domain Adaptation (VUDA) methods have been proposed to improve the robustness and transferability of video models. Despite improvements made in model robustness, these VUDA methods require access to both source data and source model parameters for adaptation, raising serious data privacy and model portability issues. To cope with the above concerns, this paper firstly formulates Black-box Video Domain Adaptation (BVDA) as a more realistic yet challenging scenario where the source video model is provided only as a black-box predictor. While a few methods for Black-box Domain Adaptation (BDA) are proposed in image domain, these methods cannot apply to video domain since video modality has more complicated temporal features that are harder to align. To address BVDA, we propose a novel Endo and eXo-TEmporal Regularized Network (EXTERN) by applying mask-to-mix strategies and video-tailored regularizations: endo-temporal regularization and exo-temporal regularization, performed across both clip and temporal features, while distilling knowledge from the predictions obtained from the black-box predictor. Empirical results demonstrate the state-of-the-art performance of EXTERN across various cross-domain closed-set and partial-set action recognition benchmarks, which even surpassed most existing video domain adaptation methods with source data accessibility.
\end{abstract}

\section{Introduction}
\label{sec:intro}

Video Unsupervised Domain Adaptation (VUDA)~\cite{chen2019temporal,choi2020shuffle,xu2021aligning,xu2021partial} aims to transfer knowledge from a labeled source video domain to an unlabeled target video domain, and has wide applications in scenarios where massive labeled video data may not be available. Despite their effectiveness in improving the robustness of video models~\cite{xu2021aligning}, current VUDA methods require access to the source video data which contains personal and private information, raising serious concerns about data privacy and model portability issues~\cite{liang2020we,liang2022dine}. The Source-Free Video Domain Adaptation~\cite{xu2022source} is subsequently formulated to learn a target model without the access of source data, but it still relies on the well-trained source model parameters which allows generative models to recover source videos~\cite{goodfellow2014generative}.

Privacy preserving is vital in applying action recognition models to real-world applications such as in smart hospitals and security surveillance where action recognition models are leveraged for anomaly behaviour recognition~\cite{sultani2018real,said2021efficient}. In these cases, current VUDA methods are totally inapplicable when sharing models across organizations due to their violation of privacy related regulations such as the European GDPR~\cite{goddard2017eu} and Singaporean PDPA~\cite{chik2013singapore}. To further cope with the video privacy issue, we therefore formulate and study a more realistic and challenging video domain adaptation scenario termed the \textbf{\textit{Black-box Video Domain Adaptation}} (BVDA) where the source video model is provided for adaptation only as a black-box predictor (e.g.,~API service). In privacy-concerned scenarios, BVDA helps to derive an accurate model in the target domain without access to both the parameters and data in the source domain.

\begin{figure}[!t]
\begin{center}
  \includegraphics[width=1.\linewidth]{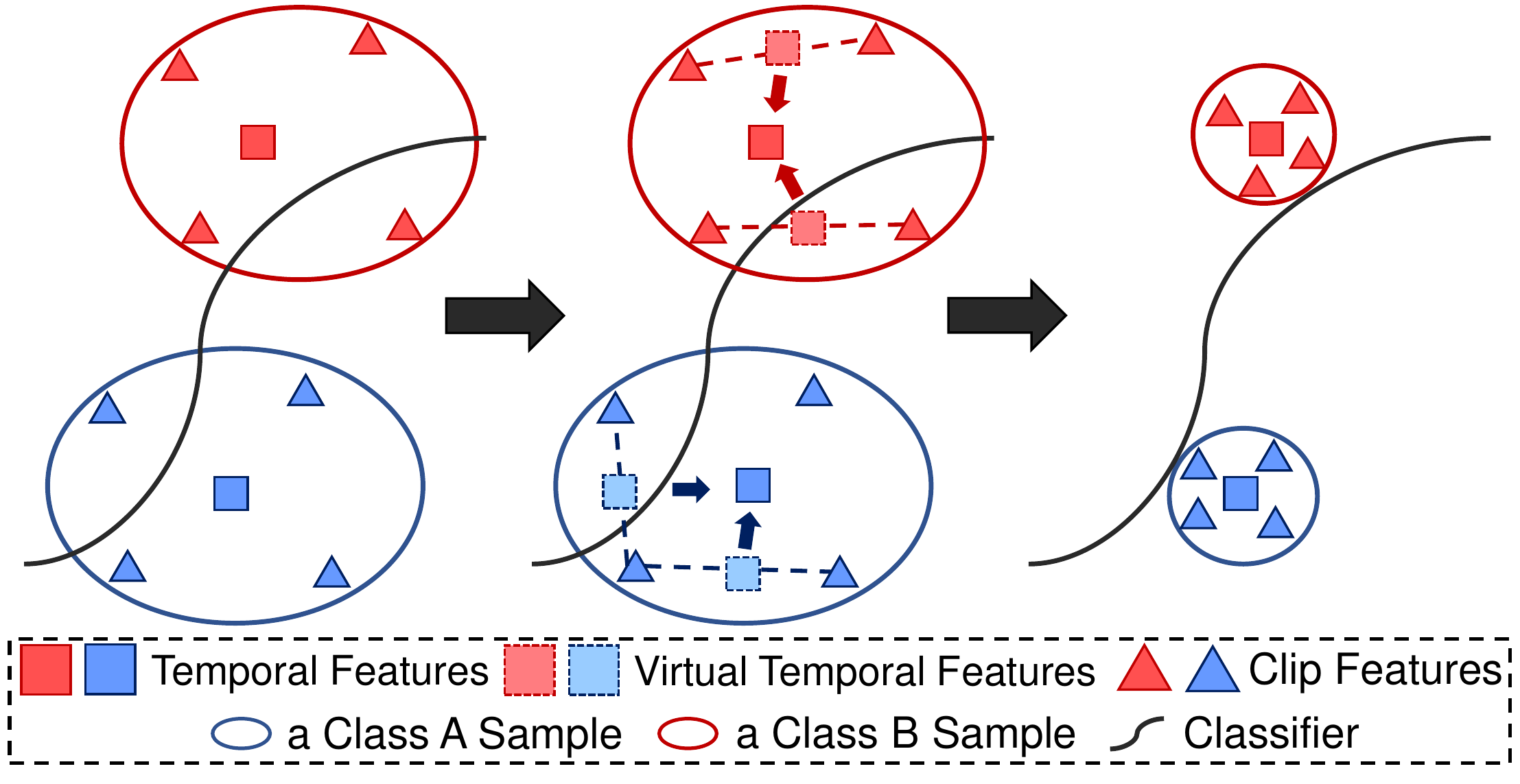}
\end{center}
\caption{Clip features of target videos may be scattered, violating both the cluster assumption and the \textit{masked temporal hypothesis}. We augment target video domain with virtual temporal features through a novel mask-to-mix strategy, and apply endo-temporal regularization. The resulting temporal features are more discriminative and complies with both the cluster assumption and the \textit{masked temporal hypothesis}.}
\label{figure:1-1-motivation}
\end{figure}

Without access to source data and model, existing VUDA methods that aim at enhancing transferability through statistical alignment (e.g., TAMAN~\cite{xu2021multi}) or adversarial alignment (e.g., TA\textsuperscript{3}N~\cite{chen2019temporal} and SAVA~\cite{choi2020shuffle}) are not applicable. There have been a few recent research efforts~\cite{liang2021source,yang2022divide} aiming at Black-box Domain Adaptation for images. One representative work is DINE~\cite{liang2022dine}, where target features are extracted by obtaining pseudo-labels from the black-box predictors while applying structural regularizations~\cite{viola1997alignment,verma2019interpolation} that encourage better model discriminability and generalization ability. However, structural regularizations of DINE are tailored for images that only contain spatial features. In comparison, characterized by the multi-modality nature, videos consist of spatial features and additional temporal information, resulting in additional challenges in aligning temporal features. As a result, solutions for images such as DINE cannot show significant improvements for the BVDA task. 
Previous studies~\cite{chen2019transferability,yang2020mind,kundu2022balancing,xu2022source} prove that improving discriminability would benefit the effectiveness of domain adaptation. Inspired by these studies, we propose to improve the discriminability of temporal features to tackle BVDA effectively when neither source data nor source model is accessible.

One common strategy to extract video temporal features is to split longer videos into shorter clips. Therefore, temporal features can be constructed explicitly with a series of clip features~\cite{wang2018temporal,zhou2018temporal}. Meanwhile, humans are capable of recognizing actions correctly even with only representative clips from videos~\cite{isik2018fast}. Intuitively, if the target model is able to perform similar to human perception and obtain discriminative features with consistent predictions given only partial clip information, the representational capacity of the target model and the discriminability of the extracted target temporal features could be improved significantly even without knowledge from the source domain. We term the above hypothesis as the \textit{masked-temporal hypothesis} as this hypothesis depicts the ideal characteristics of features obtained after certain clips are masked out. Our method is built upon this hypothesis.

To this end, we propose the \textbf{E}ndo and e\textbf{X}o-\textbf{TE}mporal \textbf{R}egularized \textbf{N}etwork (\textbf{EXTERN}) to address the BVDA task. EXTERN extracts robust temporal features in a self-supervised manner by applying both the \textit{endo-temporal regularization} and the \textit{exo-temporal regularization}, while distilling knowledge from the predictions obtained from the source predictor. Specifically, the endo-temporal regularization is designed to improve the discriminability of clip features and drive clip features towards complying with the cluster assumption~\cite{rigollet2007generalization} and the \textit{masked-temporal hypothesis} as presented in Fig.~\ref{figure:1-1-motivation}. This objective is achieved by augmenting the target video domain with virtual temporal features through a novel \textbf{mask-to-mix} strategy over clip features. Meanwhile, the exo-temporal regularization is designed to drive the proposed model to extract more stable and discriminable temporal features that are linearly smooth in-between by augmenting the target video domain with interpolated temporal features. It is remarkable that our EXTERN achieves outstanding results, outperforming most existing VUDA methods that require source data and model. This demonstrates that training the target model from scratch may help overcome the negative effect of domain shift, paving a new way for tackling VUDA.

In summary, our contributions are threefold. (i) We formulated a realistic and more challenging task, \textit{Black-box Video Domain Adaptation} (BVDA). To the best of our knowledge, this is the first work to address black-box domain adaptation for privacy-preserving and portable video model transfer. (ii) We propose EXTERN to address BVDA, which enhances discriminative temporal feature extraction through an endo-temporal regularization using a mask-to-mix strategy along with an exo-temporal regularization, driving clip features towards complying with the \textit{masked-temporal hypothesis}. (iii) Extensive experiments show the efficacy of EXTERN, achieving state-of-the-art performances across cross-domain action recognition benchmarks under closed-set and partial-set domain adaptation settings, even outperforming most existing adaptation methods.

\section{Related Work}
\label{sec:related}

\noindent
\textbf{(Video) Unsupervised Domain Adaptation ((V)UDA).}
UDA and VUDA aims to distill shared knowledge across a labeled source domain and an unlabeled target domain, which improves the robustness and transferability of deep learning models. (V)UDA methods could be broadly categorized into three categories: i) reconstruction-based methods~\cite{ghifary2016deep,yang2020label}, where domain-invariant features are extracted by encoders trained with data-reconstruction objectives, commonly constructed based on a encoder-decoder structure; ii) discrepancy-based methods~\cite{saito2018maximum,zhang2019bridging,xu2021multi}, where domain alignment is achieved by applying metric learning approaches, optimized with metric-based objectives such as MDD~\cite{zhang2019bridging}, CORAL~\cite{sun2016return}, and MMD~\cite{long2015learning}; and iii) adversarial-based methods~\cite{tzeng2017adversarial,chen2019temporal,xu2021aligning}, where methods leverage additional domain discriminators along with feature generators, trained jointly in an adversarial manner~\cite{huang2011adversarial} by minimizing adversarial losses~\cite{ganin2015unsupervised}. Compared to UDA research which is primarily focused on image-based tasks, VUDA research lags behind owing to the challenges brought by aligning temporal features. Despite the challenges, there has been a substantial increase in research for VUDA, backed by the introduction of various cross-domain closed-set or partial set video datasets~\cite{chen2019temporal,xu2021partial,xu2021multi}. Regardless of the improvements in video model robustness and transferability, VUDA approaches all require access to both source data and source model parameters during adaptation, which could raise serious privacy concerns given the amount of private information of the subjects and scenes contained in videos.

\noindent
\textbf{Black-box Domain Adaptation (BDA).}
With the increased importance of data privacy with concerns of possible data leakage through white-box attack given model parameters~\cite{zhang2020white}, there have been a few research that explores BDA with images. BDA enables image models to be adapted to the unlabeled target domain without either the source data or the source model parameters, with the source model provided only as a black-box predictor. Under such settings, BBSE~\cite{lipton2018detecting} focused on black-box label shift, and requires a hold-out source set for class confusion matrix estimation, which is commonly unavailable. More recently, LNL~\cite{zhang2021unsupervised} proposed to tackle BDA by an iterative noisy learning approach via pseudo labels that are refined with KL divergence, while DINE~\cite{liang2022dine} leveraged knowledge distillation with information maximization and structural regularizations. Despite the above recent advances, BVDA has never been explored, which is more challenging as temporal features must also be aligned. We propose to tackle BVDA by applying a temporal feature tailored endo-temporal regularization leveraging a mask-to-mix strategy, along with exo-temporal regularization.

\begin{figure*}[!ht]
\begin{center}
   \includegraphics[width=.85\linewidth]{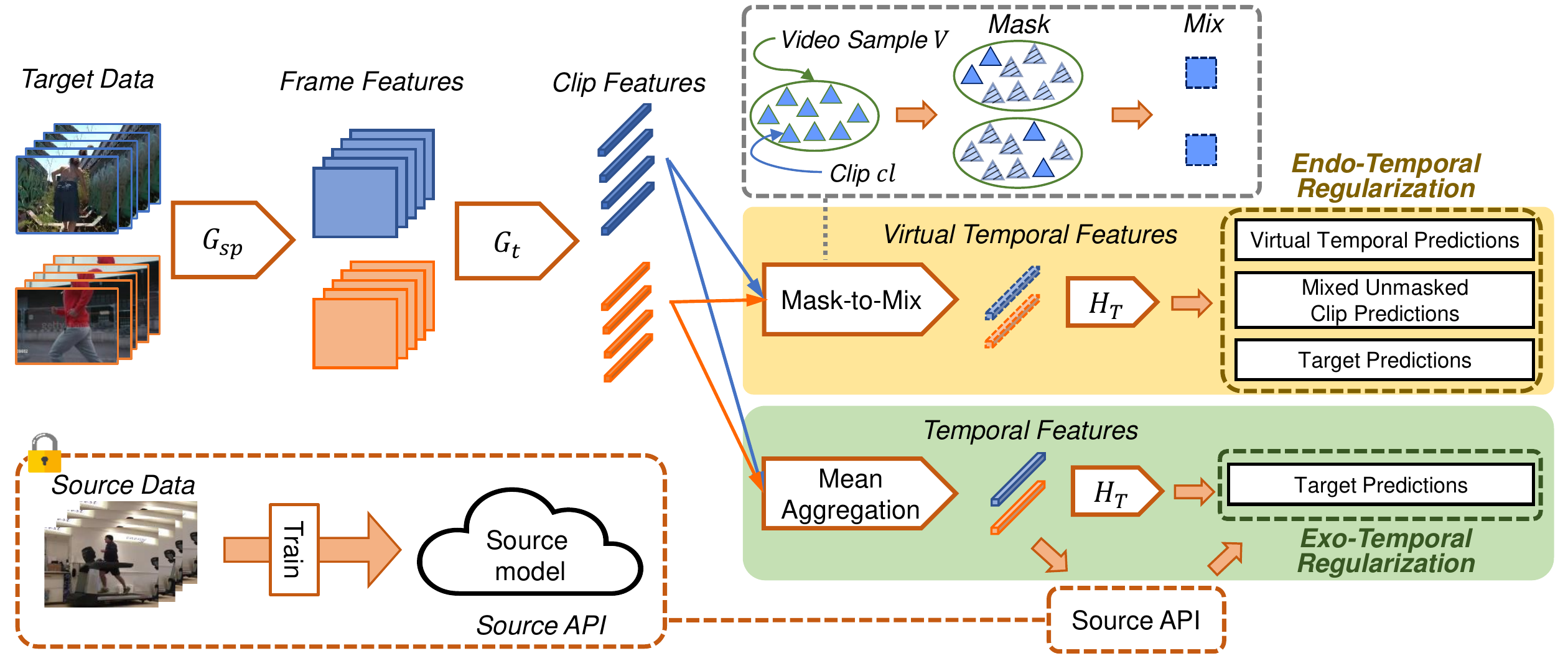}
\end{center}
\caption{An overview of the proposed EXTERN. EXTERN extracts knowledge from the black-box source predictor (i.e., Source API) through a distillation process. EXTERN further extracts temporal features in a self-supervised manner by applying both the \textit{endo-temporal regularization} and \textit{exo-temporal regularization}. To apply the \textit{endo-temporal regularization}, the virtual temporal features are constructed with a mask-to-mix strategy.}
\label{figure:3-2-structure}
\end{figure*}

\section{Methodology}
\label{sec:method}

For \textbf{\textit{Black-box Video Domain Adaptation}} (BVDA), we only have access to a black-box video predictor $H_{S}$ (i.e.,~the constraint Source API) which is trained from the labeled source video domain $\mathcal{D}_S=\{(V_{iS},y_{iS})\}^{n_S}_{i=1}$ containing $n_{S}$ videos, with $V_{iS}\in\mathcal{V}_{S}, y_{iS}\in\mathcal{Y}_{S}$. We are also given an unlabeled target video domain $\mathcal{D}_{T}=\{V_{iT}\}^{n_{T}}_{i=1}$ with $n_{T}$ i.i.d.\ videos $V_{iT}\in\mathcal{V}_{T}$. Additionally, we assume that source and target video domains share the same label space with $C$ classes, i.e.,~$\mathcal{Y}_{S}=\mathcal{Y}_{T}$ while the source and target videos follow different data distributions. Therefore, there exists a domain shift~\cite{ben2006analysis} between $\mathcal{D}_S$ and $\mathcal{D}_{T}$. The objective of BVDA is to learn a mapping model $\mathcal{V}_{T}\to\mathcal{Y}_{T}$ to perform the action recognition task on the unlabeled $\mathcal{D}_{T}$ while both $\mathcal{D}_{S}$ and the parameters of $H_{S}$ are not accessible.

Constrained by the absence of both the source data and the parameters of the source model, neither VUDA methods nor SFVDA methods could be directly applied for BVDA. To tackle BVDA, we resort to an alternative strategy where we adapt target models to the embedded semantic information of the source data resorting to the hard or soft predictions of the target domain from the black-box source predictor $\hat{\mathcal{Y}}_{T} = H_{S} (\mathcal{V}_{T})$. Essentially, for BVDA, such a strategy aims to extract effective temporal features with high discriminability and complies with the cluster assumption. We therefore propose EXTERN which drives temporal feature towards high discriminability in a self-supervised manner relying on both the \textit{endo-temporal regularization} and the \textit{exo-temporal regularization}. We first introduce the backbone structure of the target model, followed by a thorough illustration over EXTERN and its key components.

\subsection{Backbone Network}
\label{sec:method:backbone}
Videos differ from images greatly due to the existence of temporal features. A key prerequisite for the target network to be adapted to the embedded source semantic information is that its backbone could extract temporal features explicitly. An efficient and popular approach constructs the temporal features explicitly with a series of clip features, obtained through clips sampled from the corresponding videos~\cite{wang2018temporal,zhou2018temporal}. One notable example is the Temporal Relation Network (TRN)~\cite{zhou2018temporal}. TRN has been widely adopted in previous video domain adaptation tasks, such as VUDA~\cite{chen2019temporal,xu2021aligning}, PVDA~\cite{xu2021partial} and SFVDA~\cite{xu2022source} bringing state-of-the-art results, thanks to its ability in extracting accurate temporal features by reasoning over correlations between spatial representations which coincides with the process of humans recognizing actions.

Formally, we define an input target video with $k$ frames as $V_{i}=\{f_{i}^{(1)},f_{i}^{(2)},...,f_{i}^{(k)}\}$ , with $f_{i}^{(j)}$ being the spatial feature of the $j$-th frame in the $i$-th source video. The subscript for the target domain $T$ is omitted for clarity. The spatial features are extracted from the source spatial feature generator $G_{sp}$ which is formulated as a 2D-CNN (e.g., ResNet-50 or ResNet-101~\cite{he2016deep}). Subsequently, the temporal feature of $V_{i}$ is constructed by a combination of multiple clip features, obtained from the temporal feature generator $G_{t}$. More specifically, $G_{t}$ builds each clip feature with $r$ temporal-ordered sampled frames where $r\in [2,k]$. Formally, a clip feature for $V_{i}$, denoted as $cl_{i}^{(r)}$, is defined by:
\begin{equation}
\label{eqn:method:clip}
cl_{i}^{(r)} = \sum\nolimits_{m} g^{(r)}((V_{i}^{(r)})_m).
\end{equation}
Here $(V_{i}^{(r)})_m=\{f_{i}^{(a)},f_{i}^{(b)},...\}_m$ is defined as the $m$-th clip with $r$ temporal-ordered frames, where $a$ and $b$ are the frame indices. $a$ and $b$ may not be consecutive as the clip could be extracted with nonconsecutive frames, but should be both constrained within the range of $[1,k]$ with $b>a$. Eventually, the clip feature $cl_{i}^{(r)}$ is obtained by fusing all the $r$ time ordered frame-level spatial features through an integration function $g^{(r)}$, usually formulated as a Multi-Layer Perceptron (MLP). The above computation would result in a total of $(k-1)$ clips. Finally, the temporal feature of $V_{i}$, denoted as $\mathbf{t}_{i}$, is computed through the mean aggregation applied across all clip features, defined as:
\begin{equation}
\label{eqn:method:temporal}
\mathbf{t}_{i} = \frac{1}{k-1} \sum\nolimits_{r} cl_{i}^{(r)}.
\end{equation}

\subsection{Endo and Exo-Temporal Regularizations}
\label{sec:method:etr}
Current methods for BDA~\cite{liang2022dine} attempt to obtain features from the unlabeled target data with high discriminability via self-supervised learning by applying structural regularizations. However, such regularizations are only tailored for spatial features since relevant studies are only conducted over the image-based BDA tasks. Comparatively, videos contain temporal features that are constructed from a series of clip features. As depicted in Fig.~\ref{figure:1-1-motivation}, clip features of discriminative temporal features may still be scattered across the decision boundary, resulting in temporal features with indistinct semantic information and causing inferior target domain performance. The key to tackle BVDA is to improve the discriminability of clip features.

Clip features that are discriminative should meet the cluster assumption and the \textit{masked-temporal hypothesis}. Specifically, The \textit{masked-temporal hypothesis} matches the human intuition that a combination of partial clip features would result in consistent prediction as the temporal features if both the temporal features and their corresponding clip features are of high discriminability. 
In other words the clip features of high discriminability should be clustered towards the temporal feature. To achieve this, we first augment the target domain with virtual temporal features with a \textbf{mask-to-mix} strategy as depicted in Fig.~\ref{figure:3-2-structure}. We define the virtual temporal feature of $V_{i}$ as $\Tilde{\mathbf{t}}_{i}$. It is constructed by mixing a set of clip features $cl_{i}^{(r)}$, $r\in[2,k]$ selected by a random masking process. Specifically, if the temporal feature $\mathbf{t}_{i}$ is built upon a set of $(k-1)$ clips, there will be exactly $(k-3)$ clips masked randomly, leaving two randomly unmasked clips: $cl_{i}^{(r1)}$ and $cl_{i}^{(r2)}$ where $r1,r2\in[2,k]$. For each mini-batch, the selection of masked clips is random across all input videos within the mini-batch and across each epoch. This is to ensure that the virtual temporal feature is built upon all possible clip pairs across the whole training process. Different from the temporal feature which is constructed through a simple mean aggregation where all clip features are combined with equal weights, we compute the virtual temporal feature via the MixUp~\cite{zhang2018mixup} operation. Essentially, the virtual temporal feature is computed as a linear combination of the two unmasked clips assigned with random weights, defined as:
\begin{equation}
\label{eqn:method:virtual-temporal}
\begin{aligned}
\Tilde{\mathbf{t}}_{i} &= \mathrm{MixUp}_{\lambda_{v}} (cl_{i}^{(r1)}, cl_{i}^{(r2)}) \\
                       &= \lambda_{v} cl_{i}^{(r1)} + (1 - \lambda_{v}) cl_{i}^{(r2)}, 
\end{aligned}
\end{equation}
where $\lambda_{v} \in Beta(\alpha_{v}, \alpha_{v})$ is the weight assigned to $cl_{i}^{(r1)}$ sampled from a Beta distribution with $\alpha_{v}$ as the hyper-parameter. The virtual temporal feature constructed is essentially the linear intermediate representation between the unmasked clip features.

With the virtual temporal feature constructed, the \textit{masked-temporal hypothesis} is satisfied by encouraging the virtual temporal prediction to be consistent with the the prediction of the corresponding target video(or equivalently, the corresponding target temporal feature). To achieve such prediction consistency, we aim to minimize the divergence between the virtual temporal prediction and the target prediction. The virtual temporal prediction is obtained from the target predictor directly, i.e., $\Tilde{y}_{i} = H_{T} (\Tilde{\mathbf{t}}_{i})$. The target prediction is obtained by applying the target predictor to the target temporal feature, i.e., $y_{i} = H_{T} (\mathbf{t}_{i})$. The minimization of prediction divergence is formulated as:
\begin{equation}
\label{eqn:method:endo-pred-div}
\mathcal{L}_{pre} = D_{KL}(\Tilde{y}_{i} \| y_{i}),
\end{equation}
where $D_{KL}(\cdot \| \cdot)$ denotes the Kullback-Leibler divergence.

To further ensure that the temporal feature contains distinctive semantic information, the clip features should also comply with the cluster assumption. Previous studies~\cite{zhang2018mixup} suggest that the discriminability of extracted features can be improved and thus the cluster assumption is met if the feature generator behaves linearly in-between training samples. The cluster assumption of clip features is therefore complied by employing the interpolation consistency training (ICT) technique~\cite{verma2019interpolation}. Specifically, such technique encourages the virtual temporal prediction ($\Tilde{y}_{i}$) to be consistent with the mixed unmasked clip prediction. The mixed unmasked clip prediction is computed as the linear combination of the target predictions from both unmasked clips. The mixed unmasked clip prediction is computed as the linear combination of the target predictions from both unmasked clips, defined as:
\begin{equation}
\label{eqn:method:mix-clip-pred}
\begin{aligned}
{y}_{mix,i} &= \mathrm{MixUp}_{\lambda_{v}} (H_{T} (cl_{i}^{(r1)}), H_{T} (cl_{i}^{(r2)})) \\
            &= \lambda_{v} H_{T} (cl_{i}^{(r1)}) + (1 - \lambda_{v}) H_{T} (cl_{i}^{(r2)}). 
\end{aligned}
\end{equation}
Subsequently, we aim to optimize the loss function:
\begin{equation}
\label{eqn:method:endo-ict}
\mathcal{L}_{vir} = l_{ce}(\Tilde{y}_{i}, {y}_{mix,i}),
\end{equation}
where $l_{ce}$ denotes the cross-entropy loss.

Overall, the proposed \textit{endo-temporal regularization} drives the target model to obtain discriminable temporal features by extracting clip features with higher discriminability that complies with the \textit{masked-temporal hypothesis} and the cluster assumption. The \textit{endo-temporal regularization} is applied by jointly optimizing Eq.~\ref{eqn:method:endo-pred-div} and Eq.~\ref{eqn:method:endo-ict}:
\begin{equation}
\label{eqn:method:endo}
\mathcal{L}_{endo} = \mathcal{L}_{vir} + \mathcal{L}_{pre}.
\end{equation}
The implementation of the \textit{endo-temporal regularization} can be observed to be very simple, yet it brings significant improvements towards tackling BVDA, as would be presented in \cref{sec:exp}.

To further enhance the discriminability of the temporal feature, we extend the promotion of linear behavior in-between training samples towards the temporal feature. Given a pair of videos $V_{i}, V_{j}$, we employ the ICT~\cite{verma2019interpolation} across their corresponding temporal features $\mathbf{t}_{i}, \mathbf{t}_{j}$, formulating the \textit{exo-temporal regularization} term. Such operation is equivalent to augmenting the target video domain with interpolated temporal features which would drive the model towards better generalization. Formally, similar to how virtual temporal features are constructed, the interpolated temporal features of $\mathbf{t}_{i}, \mathbf{t}_{j}$ are obtained by applying MixUp~\cite{zhang2018mixup}. With $y_{i} = H_{T}(\mathbf{t}_{i})$ and $y_{j} = H_{T}(\mathbf{t}_{j})$ denoting the target predictions of $\mathbf{t}_{i}$ and $\mathbf{t}_{j}$, the exo-temporal regularization aims to optimize the loss function:
\begin{equation}
\label{eqn:method:exo}
\mathcal{L}_{exo} = l_{ce}(\mathrm{MixUp}_{\lambda_{t}} (\mathbf{t}_{i}, \mathbf{t}_{j}),\, \mathrm{MixUp}_{\lambda_{t}} (y_{i}, y_{j})),
\end{equation}
where $l_{ce}$ is the cross-entropy loss. $\mathrm{MixUp}_{\lambda_{t}}$ is defined as shown in Eq.~\ref{eqn:method:virtual-temporal} and \ref{eqn:method:mix-clip-pred}, and $\lambda_{t} \in Beta(\alpha_{t}, \alpha_{t})$ is the weight assigned to $\mathbf{t}_{i}$ with $\alpha_{t}$ as the hyper-parameter.

Our mask-to-mix strategy utilizes MixUp for the construction of virtual temporal features, which seeks to obtain consistent prediction as the temporal features such that the corresponding clip features satisfy the \textit{masked-temporal hypothesis}, which is different from existing domain adaptation works based on Mixup that regards it as a data augmentation approach~\cite{xu2020adversarial,yan2020improve,wu2020dual,panfilov2019improving}.

\subsection{Endo and eXo-TEmporal Regularized Network}
\label{sec:method:extern}
With the \textit{endo-temporal regularization} and \textit{exo-temporal regularization} terms defined, we propose the EXTERN to address BVDA leveraging on both regularizations, as depicted in Fig.~\ref{figure:3-2-structure}. EXTERN builds upon the TRN backbone structure as specified in \cref{sec:method:backbone}. 

\noindent
\textbf{Extracting Knowledge via Knowledge Distillation.} 
To extract knowledge from the black-box predictor $H_{S}$, knowledge distillation (KD)~\cite{hinton2015distilling} has proven to be an effective solution. The target model is seen as the student, and is trained to learn predictions analogous to that produced by the source model, which is seen as the teacher. However, due to domain shift between the source domain $\mathcal{D}_{S}$ and target domain $\mathcal{D}_{T}$, the output from the source model could be noisy and inaccurate. To cope with such drawback, the \textbf{adaptive label smoothing} (AdaLS) technique ~\cite{liang2022dine} with self-distillation~\cite{laine2017temporal} exponential moving average (EMA)~\cite{grebenkov2014following} update is recently proposed. Here, only the predictions of the top-$c$ classes are directly utilized while predictions of other classes are forced to a uniform distribution as:
\begin{equation}
\hat{y}_{i}^{\prime} = \mathrm{AdaLS}_{c}(\hat{y}_{i})=
\begin{cases}
  \hat{y}_{i}^{p}, & p\in \mathcal{T}_{i}^{c} \\
  \frac{1-\sum\nolimits_{q\in \mathcal{T}_{i}^{c}} \hat{y}_{i}^{q}}{C-c}, & \text{otherwise,}
\end{cases}
\end{equation}
where $\hat{y}_{i}\in\hat{\mathcal{Y}}$ is the prediction of the target video $V_{i}$ obtained from the black-box source predictor $H_{S}$ (i.e., the teacher prediction), while $y_{i}^{p}$ denotes the prediction of the $p$-th class and $\mathcal{T}_{i}^{c}$ is the class index set of the top-$c$ predictions for input $V_{i}$. The teacher prediction is further dynamically updated per training epoch to maintain a EMA prediction. We apply AdaLS with EMA to reduce noisy information by focusing only the top-predicted classes. 
Extracting source knowledge is eventually achieved by optimizing:
\begin{equation}
\label{eqn:method:kd}
\mathcal{L}_{kd} = \mathbb{E}_{V_{i}\in \mathbf{D}_{T}} D_{KL}(\hat{y}_{i}^{\prime} \| y_{i}).
\end{equation}

\noindent
\textbf{Learning Adaptive Clip Weights.}
Previous research~\cite{nguyen2020domain,chen2019temporal} shows that features with lower prediction uncertainty would possess higher discriminability. Therefore, to better aggregate the clip features for the temporal feature, we assign a \textbf{clip weight} to each clip feature by attending to clip features with lower prediction uncertainty. Specifically, the clip weight is defined as the additive inverse of the target predictions of the corresponding clip, computed as:
\begin{equation}
\label{eqn:method:clip-weight}
w_{cl_{i}^{(r)}} = 1 - h(H_{T} (cl_{i}^{(r)})),
\end{equation}
where the constant 1 is added as a residual connection for more stable optimization. Subsequently, the temporal feature is obtained as the weighted aggregation of all clip features, with Eq.~\ref{eqn:method:temporal} modified as:
\begin{equation}
\label{eqn:method:temporal-weighted}
\mathbf{t}_{i} = \frac{1}{k-1} \sum\nolimits_{r} w_{cl_{i}^{(r)}}\, cl_{i}^{(r)}.
\end{equation}

\noindent
\textbf{Information Maximization.} 
Inspired by prior works in BDA~\cite{liang2022dine,yang2022divide}, we maximize the \textbf{mutual information} (MI) to encourage diversity among target predictions and to promote their individual certainty:
\begin{equation}
\label{eqn:method:mi}
\begin{aligned}
\mathcal{L}_{mi} &= h(\mathbb{E}_{V_{i}\in \mathbf{D}_{T}} y_{i}) - \mathbb{E}_{V_{i}\in \mathbf{D}_{T}} h(y_{i}),
\end{aligned}
\end{equation}
where $y_{i} = H_{T}(G_{sp}(G_{t}(V_{i}))) = H_{T}(\mathbf{t}_{i})$ is the target prediction for input $V_{i}$ and $h(y_{i}) = - \sum\nolimits_{c=1}^{C} y_{i}^{c}\,\log\,y_{i}^{c}$ is the conditional entropy function. Maximizing MI could \textbf{marginally} improve the performances for BVDA, as would be shown later in the ablation studies (\cref{sec:exp:analysis}).

\begin{table*}[!ht]
\center
\resizebox{.9\linewidth}{!}{\noindent
\begin{tabular}{l|l|cc|ccc|ccccccc}
\hline
\hline
  \multirow{2}{*}{Methods} &
  \multirow{2}{*}{Publication} &
  \multicolumn{2}{c|}{\textbf{Privacy}} &
  \multicolumn{3}{c|}{\textbf{UCF-HMDB\textsubscript{\textit{full}}}} &
  \multicolumn{7}{c}{\textbf{Sports-DA}} \Tstrut\Bstrut\\
\cline{3-14}
& & Data & Model & U101$\to$H51 & H51$\to$U101 & Avg. & K600$\to$U101 & K600$\to$S1M & S1M$\to$U101 & S1M$\to$K600 & U101$\to$K600 & U101$\to$S1M & Avg.\\
\hline
TRN~\cite{zhou2018temporal}& ECCV-18 & - & -
& 76.11 & 78.97 & 77.54 & 90.25 & 71.16 & 88.95 & 73.90 & 62.73 & 49.74 & 72.79 \\
\hline
LNL~\cite{zhang2021unsupervised} & - & \cmark & \cmark
& 75.78 & 78.92 & 77.35 & 82.37 & 68.44 & 82.11 & 73.11 & 59.03 & 54.84 & 69.98 \\
HD-SHOT~\cite{liang2021source} & TPAMI(21') & \cmark & \cmark
& 77.86 & 80.39 & 79.13 & 87.08 & 69.75 & 81.59 & 72.11 & 65.63 & 60.49 & 72.78 \\
SD-SHOT~\cite{liang2021source} & TPAMI(21') & \cmark & \cmark
& 79.29 & 82.22 & 80.76 & 85.39 & 68.07 & 83.58 & 74.80 & 63.94 & 60.75 & 72.75 \\
DINE~\cite{liang2022dine} & CVPR-21 & \cmark & \cmark
& 81.39 & 87.57 & 84.48 & 91.60 & 72.11 & 86.54 & 77.59 & 76.22 & 66.95 & 78.50 \\
\hline
EXTERN & - & \cmark & \cmark
& \textbf{88.89} & \textbf{91.95} & \textbf{90.42} & \textbf{93.77} & \textbf{73.79} & \textbf{95.42} & \textbf{82.16} & \textbf{81.19} & \textbf{72.74} & \;\textbf{83.18} \Tstrut \Bstrut\\
\hline
TA\textsuperscript{3}N~\cite{chen2019temporal} & ICCV-19 & \xmark & \xmark
& 77.70 & 85.37 & 81.54 & 90.28 & 68.57 & 92.97 & 72.65 & 63.63 & 54.06 & 73.70 \Tstrut\\
DANN~\cite{ganin2015unsupervised} & ICML-15 & \xmark & \xmark
& 78.63 & 90.29 & 84.46 & 87.97 & 75.05 & 85.75 & 73.40 & 65.88 & 55.08 & 73.85 \\
MK-MMD~\cite{long2015learning} & ICML-15 & \xmark & \xmark
& 77.99 & 86.18 & 82.09 & 90.16 & 67.95 & 90.95 & 73.58 & 66.10 & 55.58 & 74.05 \\
SAVA~\cite{choi2020shuffle} & ECCV-20 & \xmark & \xmark
& 78.56 & 89.28 & 83.92 & 97.33 & 75.76 & 91.20 & 75.28 & 58.17 & 51.33 & 74.85 \\
SHOT~\cite{liang2021source} & TPAMI(21') & \cmark & \xmark
& 77.44 & 86.77 & 82.10 & 91.91 & 72.44 & 91.95 & 75.57 & 67.81 & 52.11 & 75.30 \\
ACAN~\cite{xu2021aligning} & - & \xmark & \xmark
& 84.04 & 93.78 & 88.91 & 94.70 & 76.69 & 92.32 & 77.69 & 62.50 & 52.38 & 76.05 \\
ATCoN~\cite{xu2022source} & ECCV-22 & \cmark & \xmark
& 83.21 & 91.07 & 87.14 & 97.59 & 77.56 & 94.36 & 80.32 & 67.20 & 55.17 & 78.70 \\
\hline
\hline
\end{tabular}
}
\caption{Results for BVDA on UCF-HMDB\textsubscript{\textit{full}} and Sports-DA for closed-set video domain adaptation.}
\label{table:4-1-sota-vanilla-1}
\end{table*}

\noindent
\textbf{Overall Objective.}
Summarizing all the loss functions as presented in Eqs.~(\ref{eqn:method:endo}, \ref{eqn:method:exo}, \ref{eqn:method:kd}, \ref{eqn:method:mi}), the overall optimization objective of EXTERN is expressed as:
\begin{equation}
\label{eqn:method:overall-obj}
\mathcal{L} = \mathcal{L}_{kd} + \beta_{reg}(\mathcal{L}_{endo} + \mathcal{L}_{exo}) - \mathcal{L}_{mi},
\end{equation}
where $\beta_{reg}$ is the hyper-parameter that controls the strength of the regularizations and is empirically set to 1. We refer to the settings of DINE~\cite{liang2022dine} by setting $\alpha_{t}$ as 0.3 and $c$ as 3.

\begin{table*}[!t]
\center
\resizebox{.95\linewidth}{!}{\noindent
\begin{tabular}{l|cc|ccccccccccccc}
\hline
\hline
  \multirow{2}{*}{Methods} &
  \multicolumn{2}{c|}{\textbf{Privacy}} &
  \multicolumn{13}{c}{\textbf{Daily-DA}} \Tstrut\Bstrut\\
\cline{2-16}
& Data & Model & K600$\to$A11 & K600$\to$H51 & K600$\to$MIT & MIT$\to$A11 & MIT$\to$H51 & MIT$\to$K600 & H51$\to$A11 & H51$\to$MIT & H51$\to$K600 & A11$\to$H51 & A11$\to$MIT & A11$\to$K600 & Avg. \Tstrut\\
\hline
TRN~\cite{zhou2018temporal} & - & -
& 25.91 & 37.50 & 31.25 & 20.25 & 45.83 & 61.66 & 16.99 & 33.25 & 43.45 & 20.42 & 13.25 & 21.66 & 30.95 \\
\hline
LNL~\cite{zhang2021unsupervised} & \cmark & \cmark
& 20.75 & 49.38 & 32.25 & 15.51 & 41.52 & 55.96 & 16.80 & 31.75 & 41.34 & 20.04 & 14.00 & 35.85 & 31.26 \\
HD-SHOT~\cite{liang2021source} & \cmark & \cmark
& 15.84 & 46.87 & 32.50 & 16.26 & 39.14 & 56.52 & 15.87 & 31.00 & 43.12 & 23.28 & 15.25 & 42.60 & 31.52 \\
SD-SHOT~\cite{liang2021source} & \cmark & \cmark
& 17.02 & 47.92 & 33.25 & 16.56 & 41.07 & 58.16 & 16.17 & 32.50 & 46.96 & 24.49 & 16.00 & 45.57 & 32.97 \\
DINE~\cite{liang2022dine} & \cmark & \cmark
& 19.47 & 50.83 & 34.50 & 14.28 & 49.17 & 64.00 & 23.51 & 38.75 & 51.17 & 21.25 & 17.75 & 47.03 & 35.98 \\
\hline
EXTERN & \cmark & \cmark
& \textbf{23.97} &  \textbf{55.83} &  \textbf{35.25} &  \textbf{18.15} &  \textbf{53.75} &  \textbf{68.14} &  \textbf{26.22} &  \textbf{40.75} &  \textbf{57.66} &  \textbf{26.25} &  \textbf{18.25} &  \textbf{51.45} &  \,\textbf{39.64} \Tstrut \Bstrut\\
\hline
TA\textsuperscript{3}N~\cite{chen2019temporal} & \xmark & \xmark
& 23.51 & 36.17 & 31.75 & 18.94 & 43.77 & 57.19 & 16.58 & 28.75 & 40.38 & 17.81 & 14.00 & 22.04 & 29.24 \Tstrut\\
DANN~\cite{ganin2015unsupervised} & \xmark & \xmark
& 25.30 & 38.34 & 23.25 & 20.71 & 45.30 & 61.86 & 16.86 & 35.25 & 40.26 & 24.46 & 19.00 & 27.38 & 31.50 \\
MK-MMD~\cite{long2015learning} & \xmark & \xmark
& 25.88 & 37.06 & 25.75 & 19.09 & 52.71 & 61.57 & 24.16 & 30.75 & 35.58 & 22.81 & 17.25 & 26.40 & 31.58 \\
SAVA~\cite{choi2020shuffle} & \xmark & \xmark
& 26.33 & 38.29 & 32.00 & 20.61 & 46.50 & 62.64 & 21.30 & 34.00 & 44.38 & 23.74 & 13.50 & 22.08 & 32.11 \\
SHOT~\cite{liang2021source} & \cmark & \xmark
& 18.37 & 48.40 & 34.50 & 13.88 & 38.33 & 53.73 & 22.05 & 29.00 & 47.92 & 31.93 & 16.50 & 39.52 & 32.84 \\
ACAN~\cite{xu2021aligning} & \xmark & \xmark
& 27.08 & 42.39 & 33.50 & 21.17 & 47.97 & 63.88 & 21.81 & 34.75 & 45.79 & 25.35 & 15.00 & 31.73 & 34.20 \\
ATCoN~\cite{xu2022source} & \cmark & \xmark
& 22.55 & 53.32 & 35.00 & 24.73 & 52.50 & 65.90 & 25.28 & 36.75 & 53.51 & 32.44 & 17.00 & 43.45 & 38.54 \\
\hline
\hline
\end{tabular}
}
\caption{Results for BVDA on Daily-DA for closed-set video domain adaptation.}
\label{table:4-2-sota-vanilla-2}
\end{table*}

\section{Experiments}
\label{sec:exp}

In this section, we evaluate our proposed EXTERN across a variety of cross-domain action recognition benchmarks, covering a wide range of cross-domain scenarios. We demonstrate exceptional performances on all benchmarks. Moreover, thorough ablation studies and analysis of EXTERN are performed to further justify the design of EXTERN.

\subsection{Experimental Settings}
\label{sec:exp:setting}
\noindent
\textbf{Datasets.}
We evaluate EXTERN on three benchmarks: UCF-HMDB\textsubscript{\textit{full}}~\cite{chen2019temporal}, Sports-DA~\cite{xu2021multi} and Daily-DA~\cite{xu2021multi}. \textbf{UCF-HMDB\textsubscript{\textit{full}}} is one of the most common benchmarks for VUDA, and is constructed from UCF101 (U101)~\cite{soomro2012ucf101} and HMDB51 (H51)~\cite{kuehne2011hmdb}, each corresponding to a separate domain. \textbf{Sports-DA} is a large-scale benchmark with three domains, built from UCF101, Sports-1M (S1M)~\cite{karpathy2014large}, and Kinetics (K600). \textbf{Daily-DA} incorporates both normal and low-illumination videos with four domains, built from ARID (A11)~\cite{xu2021arid} (a low-illumination video dataset), HMDB51, Moments-in-Time (MIT)~\cite{monfort2019moments}, and Kinetics~\cite{kay2017kinetics}. The distant domain shift due to immense illumination difference render it more challenging.

\noindent
\textbf{Implementation.}
We implement our method with the PyTorch~\cite{paszke2019pytorch} library. We adopt the TRN~\cite{zhou2018temporal} built upon ResNet-50~\cite{he2016deep} as the backbone for video feature extraction, pre-trained on ImageNet~\cite{deng2009imagenet}. Hyper-parameters $\alpha_{v}=0.3$ and $\beta_{reg}=1.0$ are empirically set and fixed in this paper. 

\noindent
\textbf{Baselines.}
We compare EXTERN with state-of-the-art BDA approaches as well as competitive UDA/VDA approaches. For fair comparisons, all compared methods are re-implemented with the exact same backbone as EXTERN. Specifically, for BDA approaches, \textbf{LNL}~\cite{zhang2021unsupervised} is a noisy label learning method where pseudo labels are refined with KL divergence and leveraged for iterative network training. \textbf{HD-SHOT} and \textbf{SD-SHOT} obtain the model through self-training and apply SHOT~\cite{liang2021source} by employing a cross-entropy loss and weighted cross-entropy loss respectively. We also compare with methods including: DINE~\cite{liang2022dine}, DANN~\cite{ganin2015unsupervised}, MK-MMD~\cite{long2015learning}, TA\textsuperscript{3}N~\cite{chen2019temporal}, SAVA~\cite{choi2020shuffle}, ACAN~\cite{xu2021aligning}, SHOT~\cite{liang2021source}, ATCoN~\cite{xu2022source}, BA\textsuperscript{3}US~\cite{liang2020balanced}, PADA~\cite{cao2018partial} and PATAN~\cite{xu2021partial}. We report the average accuracies over five runs with identical settings.

\begin{table*}[!ht]
\center
\resizebox{.8\linewidth}{!}{\noindent
\begin{tabular}{l|l|cc|ccc|ccc|ccc}
\hline
\hline
  \multirow{2}{*}{Methods} &
  \multirow{2}{*}{Publication} &
  \multicolumn{2}{c|}{\textbf{Privacy}} &
  \multicolumn{3}{c|}{UCF-HMDB\textsubscript{\textit{partial}}} &
  \multicolumn{3}{c|}{HMDB-ARID\textsubscript{\textit{partial}}} &
  \multicolumn{3}{c}{MiniKinetics-UCF} \Tstrut\Bstrut\\
\cline{3-13}
& & Data & Model & U-14$\to$H-7 & H-14$\to$U-7 & Avg. & H-10$\to$A-5 & A-10$\to$H-5 & Avg. & M-45$\to$U-18 & U-45$\to$M-18 & Avg. \\
\hline
TRN~\cite{zhou2018temporal} & ECCV-18 & - & -
& 59.05 & 82.33 & 70.69 & 21.54 & 29.33 & 25.44 & 64.30 & 87.56 & 75.93 \\
LNL~\cite{zhang2021unsupervised} & - & \cmark & \cmark
& 56.79 & 80.94 & 68.86 & 22.23 & 26.57 & 24.40 & 61.40 & 85.92 & 73.66 \\
HD-SHOT~\cite{liang2021source} & TPAMI(21') & \cmark & \cmark
& 56.41 & 80.62 & 68.51 & 23.30 & 26.84 & 25.07 & 59.95 & 89.62 & 74.78 \\
SD-SHOT~\cite{liang2021source} & TPAMI(21') & \cmark & \cmark
& 61.52 & 82.42 & 71.97 & \textbf{23.74} & 25.62 & 24.68 & 61.07 & 88.79 & 74.93 \\
DINE~\cite{liang2022dine} & CVPR-21 & \cmark & \cmark
& 66.19 & 83.84 & 75.01 & 17.69 & 17.33 & 17.51 & 68.79 & 93.56 & 81.18 \\
\hline
EXTERN & - & \cmark & \cmark
& \textbf{71.43} & \textbf{90.60} & \textbf{81.02} & 23.08 & \textbf{38.67} & \textbf{30.87} & \textbf{75.89} & \textbf{96.49} & \;\textbf{86.19} \Tstrut \Bstrut\\
\hline
TA\textsuperscript{3}N~\cite{chen2019temporal} & ICCV-19 & \xmark & \xmark
& 50.99 & 73.70 & 62.35 & 20.95 & 27.08 & 24.02 & 63.24 & 92.14 & 77.69 \Tstrut\\
DANN~\cite{ganin2015unsupervised} & ICML-15 & \xmark & \xmark
& 61.56 & 77.63 & 69.59 & 22.73 & 19.54 & 21.13 & 62.06 & 93.04 & 77.55 \\
MK-MMD~\cite{long2015learning} & ICML-15 & \xmark & \xmark
& 59.16 & 82.25 & 70.70 & 22.31 & 25.79 & 24.05 & 69.26 & 88.69 & 78.98 \\
SAVA~\cite{choi2020shuffle} & ECCV-20 & \xmark & \xmark
& 54.74 & 83.41 & 69.08 & 25.27 & 27.94 & 26.61 & 66.49 & 90.31 & 78.40 \\
PADA~\cite{cao2018partial} & ECCV-18 & \xmark & \xmark
& 68.37 & 85.86 & 77.11 & 21.28 & 32.60 & 26.94 & 72.72 & 91.62 & 82.17 \\
BA\textsuperscript{3}US~\cite{liang2020balanced} & ECCV-20 & \xmark & \xmark
& 71.85 & 88.41 & 80.13 & 26.81 & 32.20 & 29.51 & 76.41 & 95.44 & 85.93 \\
PATAN~\cite{xu2021partial} & ICCV-21 & \xmark & \xmark
& 73.60 & 91.85 & 82.72 & 30.34 & 35.51 & 32.93 & 77.31 & 96.50 & 86.90 \\
\hline
\hline
\end{tabular}
}
\caption{Results for BVDA on UCF-HMDB\textsubscript{\textit{partial}}, HMDB-ARID\textsubscript{\textit{partial}} and MiniKinetics-UCF for partial-set video domain adaptation.}
\label{table:4-3-sota-partial}
\end{table*}

\begin{table*}[!ht]
\center
\resizebox{.6\linewidth}{!}{\noindent
\begin{tabular}{l|c|cc|cc|c}
\hline
\hline
  \multirow{2}{*}{Methods} &
  \multirow{2}{*}{Clip Weights} &
  \multicolumn{2}{c|}{UCF-HMDB\textsubscript{\textit{full}}} &
  \multicolumn{2}{c|}{UCF-HMDB\textsubscript{\textit{partial}}} &
  \multirow{2}{*}{Avg.}
  \Tstrut\Bstrut\\
\cline{3-6}
&  & U101$\to$H51 & H51$\to$U101 & U-14$\to$H-7 & H-14$\to$U-7 \Tstrut\\
\hline
\multirow{2}{*}{EXTERN (w/o $\mathcal{L}_{endo}$)} 
& \xmark & 78.25 & 85.64 & 60.72 & 81.09 & \;76.42 \Tstrut\\
& \cmark & 80.08 & 86.34 & 61.91 & 82.71 & 77.76 \\
\multirow{2}{*}{EXTERN (w/o $\mathcal{L}_{exo}$)} 
& \xmark & 84.65 & 90.11 & 66.62 & 87.59 & \;82.24 \Tstrut\\
& \cmark & 85.83 & 90.37 & 67.71 & 88.91 & 83.21 \\
\multirow{2}{*}{EXTERN (w/o $\mathcal{L}_{vir}$)} 
& \xmark & 82.66 & 89.49 & 64.52 & 86.47 & \;80.78 \Tstrut\\
& \cmark & 83.89 & 89.93 & 65.24 & 88.16 & 81.80 \\
\multirow{2}{*}{EXTERN (w/o $\mathcal{L}_{pre}$)} 
& \xmark & 86.47 & 90.63 & 68.11 & 88.23 & \;83.36 \Tstrut\\
& \cmark & 87.22 & 90.89 & 69.05 & 89.47 & 84.16 \\
\multirow{2}{*}{EXTERN (w/o $\mathcal{L}_{mi}$)} 
& \xmark & 87.56 & 91.02 & 70.07 & 89.69 & \;84.75 \Tstrut\\
& \cmark & 88.43 & 91.64 & 71.06 & 90.34 & 85.36 \\
\hline
\multirow{2}{*}{EXTERN} 
& \xmark & 87.92 & 91.33 & 70.38 & 90.08 & \;84.93 \Tstrut\\
& \cmark & \textbf{88.89} & \textbf{91.95} & \textbf{71.43} & \textbf{90.60} & \;\textbf{85.72} \Bstrut\\
\hline
\hline
\end{tabular}
}
\caption{Ablation studies of learning objectives and clip weights on UCF-HMDB\textsubscript{\textit{full}} and UCF-HMDB\textsubscript{\textit{partial}}.}
\label{table:4-4-ablation}
\end{table*}

\subsection{Overall Results and Comparisons}
\label{sec:exp:compare}
\noindent
\textbf{Closed-set domain adaptation.}
We show the results on UCF-HMDB\textsubscript{\textit{full}} and Sports-DA in \cref{table:4-1-sota-vanilla-1}, and results on Daily-DA in \cref{table:4-2-sota-vanilla-2}. Our proposed EXTERN achieved state-of-the-art results across all the three cross-domain benchmarks. On average, EXTERN outperforms all BDA approaches designed for image-based DA tasks (i.e., LNL, HD/SD-SHOT and DINE), outperforming the best method by a relative $7.0\%$, $6.0\%$ and $10.2\%$ respectively. This justifies the effectiveness of the designed regularizations tailored for temporal features whose discriminability relies on clip features complying with the \textit{masked-temporal hypothesis} and the cluster assumption. It could also be observed that the prior BDA approaches may fail to tackle BVDA well, with at least one task of Sports-DA and Daily-DA benchmarks showing inferior performance to that of the source-only model. Prior BDA approaches focused solely on spatial features, and may not obtain clip features that meet the \textit{masked-temporal hypothesis} and temporal feature with distinct semantic information, resulting in negative impacts compared to the source-only baseline. Notably, EXTERN even outperforms various VUDA approaches~\cite{chen2019temporal,choi2020shuffle,xu2021aligning} with source data accessibility. Since EXTERN performs adaptation based solely on prediction results, it is found that EXTERN would not be affected by noise contained within the source data or source model, generating superior adaptation results. This shows that training a target model from scratch with strong regularizations while adapting solely with source predictions can be as effective as data-based domain alignment techniques.

\noindent
\textbf{Partial-set domain adaptation.}
Apart from closed-set video domain adaptation, we further demonstrate the generalization ability of our proposed EXTERN by evaluating on partial-set video domain adaptation (PVDA) tasks. To achieve this, we follow~\cite{xu2021partial} and leverage on three other benchmarks: UCF-HMDB\textsubscript{\textit{partial}}, HMDB-ARID\textsubscript{\textit{partial}} and MiniKinetics-UCF. 

\textbf{UCF-HMDB\textsubscript{\textit{partial}}} is built from UCF101 and HMDB51 from 14 overlapping categories and contains two PVDA tasks: \textbf{U-14}$\to$\textbf{H-7} and \textbf{H-14}$\to$\textbf{U-7}. \textbf{HMDB-ARID\textsubscript{\textit{partial}}} is built from HMDB51 and ARID, which is more challenging thanks to the distant domain shift. The dataset is collected from 10 and contains two PVDA tasks: \textbf{H-10}$\to$\textbf{A-5} and \textbf{A-10}$\to$\textbf{H-5}. \textbf{MiniKinetics-UCF} is a large-scale dataset built from MiniKinetics~\cite{xie2017rethinking} and UCF101 containing 45 overlapping categories, also containing two PVDA tasks: \textbf{M-45}$\to$\textbf{U-18} and \textbf{U-45}$\to$\textbf{M-18}.

\begin{figure*}[!htbp]
\begin{center}
  \includegraphics[width=1.\linewidth]{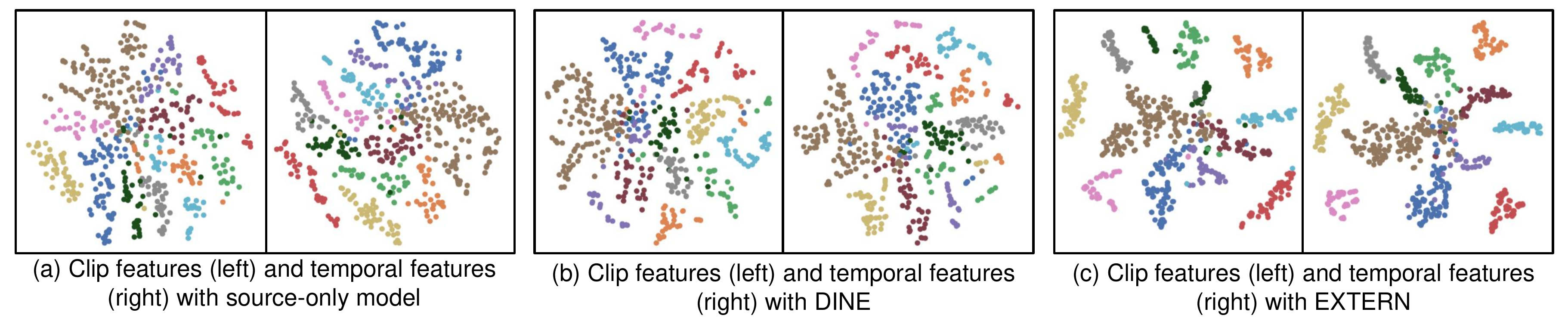}
\end{center}
\caption{t-SNE~\cite{van2008visualizing} visualizations of clip features and temporal features extracted by (a) source-only model, (b) DINE~\cite{liang2022dine}, and (c) EXTERN, with class information. Different colors denotes different classes.}
\label{figure:4-3-tsne}
\end{figure*}

\begin{figure}[!htbp]
\begin{center}
  \includegraphics[width=1.\linewidth]{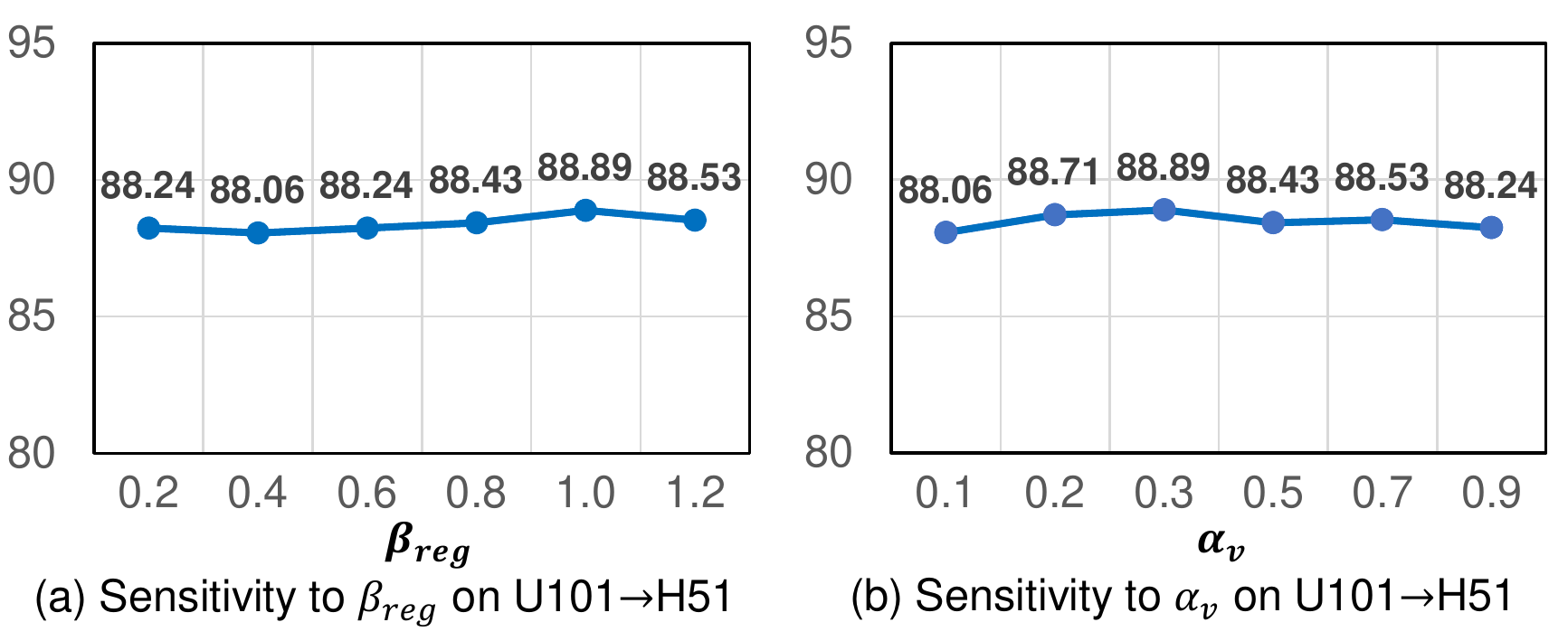}
\end{center}
\caption{Sensitivity of $\beta_{reg}$ and $\alpha_{v}$ on UCF-HMDB\textsubscript{\textit{full}}.}
\label{figure:4-4-hyperparameter}
\end{figure}

The results for partial-set cross-domain action recognition is presented in \cref{table:4-3-sota-partial}. Partial-set is more challenging due to the asymmetric label spaces with the existence of ``source-only'' classes, causing negative transfer~\cite{cao2018partial,xu2021partial}. Negative transfer affects all previous BDA approaches, where all approaches will under-perform against the source-only baseline in at least one benchmark. Despite such challenge, EXTERN still achieves outstanding results, outperforming the best BDA approach by a relative $8.0\%$, $23.1\%$ and $6.2\%$ on the three benchmarks respectively. EXTERN also surpasses several PDA approaches even though EXTERN is not specifically catered for label shift.

\subsection{Analysis}
\label{sec:exp:analysis}
\noindent
\textbf{Ablation studies.}
To gain deeper understanding of the effectiveness of EXTERN while justifying its design, we perform detailed ablation studies as shown in \cref{table:4-4-ablation}. Specifically, the ablations studies explore EXTERN from three perspectives: firstly, the effects of key learning objectives that constitute the regularization terms; secondly the effects of assigning clip weights; and lastly the effects of applying information maximization. Ablation studies are conducted on UCF-HMDB\textsubscript{\textit{full}} and UCF-HMDB\textsubscript{\textit{partial}} with the same TRN backbone as previous experiments. With any of the five learning objectives (i.e., $\mathcal{L}_{endo}$, $\mathcal{L}_{exo}$, $\mathcal{L}_{vir}$, $\mathcal{L}_{pre}$ and $\mathcal{L}_{mi}$) removed, the respective variants of EXTERN performs inferior to the original EXTERN, thus justifying that the designed learning objectives complement each other. It could be observed that by applying the \textit{endo-temporal regularization} alone (variant EXTERN (w/o $\mathcal{L}_{exo}$)), EXTERN could outperform all prior BDA approaches. This further prove the effectiveness of \textit{endo-temporal regularization} since this regularization is tailored to temporal features. Meanwhile, the improvements brought by constructing temporal features attentively with clip weight is consistent among all EXTERN variants. However, such improvement is marginal compared to that brought by the temporal feature-tailored regularization. Similarly, the improvements from applying information maximization is also consistent, yet even more negligible than that brought by the application of clip weight.

\noindent
\textbf{Hyper-parameter Sensitivity.}
We focus on studying the hyper-parameter sensitivity of $\beta_{reg}$ which controls the strength of the regularizations and $\alpha_{v}$ which relates to the construction of virtual temporal feature $\Tilde{\mathbf{t}}_{i}$.
Here $\beta_{reg}$ is in the range of 0.2 to 1.2 and $\alpha_{v}$ is in the range of 0.1 to 0.9. As shown in Fig. \ref{figure:4-4-hyperparameter}, the results of EXTERN falls within a marginal $0.83\%$, ranging from $88.06\%$ to $88.89\%$, with the best results obtained at $\alpha_{v}=0.3$ and $\beta_{reg}=1.0$. The minimal variations shows that the performance of EXTERN is robust to both hyper-parameters. Meanwhile, despite the slight variations, EXTERN maintains the best results with all the hyper-parameter settings.

\noindent
\textbf{Feature Visualization.}
We further understand the characteristics of EXTERN by plotting the t-SNE embeddings~\cite{van2008visualizing} of both the clip features and temporal features extracted by the source-only model, DINE and EXTERN for H51$\to$U101, as shown in Fig.~\ref{figure:4-3-tsne}. It is clearly observed that both the clip features and temporal features from EXTERN are more clustered and discriminable, justifying that the applied regularizations promotes higher discriminability and better compliance to the cluster assumption. We can also observe that the distribution of clip features is more similar to the distribution of temporal features with EXTERN. This intuitively proves that EXTERN drives clip features towards satisfying the \textit{masked-temporal hypothesis} where clip features are aligned towards the temporal features, and ensures that the temporal features contain distinct semantic information with high discriminability.

\section{Conclusion}
\label{sec:conclusion}

In this work, we pioneer in formulating and exploring the realistic yet more challenging task of \textit{Black-box Video Domain Adaptation} (BVDA) for privacy-preserving and portable video model transfer. To tackle BVDA, we propose EXTERN which obtains effective and discriminative temporal features by driving clip features to satisfy the \textit{masked-temporal hypothesis} and comply with the cluster assumption. This is achieved by applying a novel \textit{endo-temporal regularization} following a mask-to-mix strategy, along with an exo-temporal regularization. EXTERN drives the temporal features to contain distinct semantic information with high discriminability. Extensive empirical results across cross-domain action recognition benchmarks under both closed-set and partial-set domain adaptation settings with detailed analysis justifies the efficacy of EXTERN in tackling BVDA. We believe such the superior performance of EXTERN whose adaptation is based solely on source prediction without access to source data and model could pave a new way for tackling video domain adaptation.

{\small
\bibliographystyle{ieee_fullname}
\bibliography{cvpr}
}

\end{document}